\documentclass[conference]{IEEEtran}

\usepackage{amsmath}
\usepackage{graphicx}
\usepackage{amssymb}
\usepackage{cite}
\usepackage{hyperref}
\usepackage{makecell}
\usepackage{multirow}
\usepackage[capitalise]{cleveref}
\usepackage{textgreek}
\usepackage{newtxtext,newtxmath}
\usepackage[utf8]{inputenc}
\usepackage[letterpaper,left=0.64in,right=0.64in,top=0.76in,bottom=1.013in]{geometry}

\usepackage{subcaption}    
\usepackage{caption}       % For advanced caption control
\usepackage{float}         % To force figure placement if needed
\usepackage{amsmath}       % For mathematical symbols
\usepackage{booktabs}      % For better tables

\title{Domain Adaptive Skin Lesion Classification via Conformal Ensemble of Vision Transformers}

\author{
    \IEEEauthorblockN{Mehran Zoravar}
    \IEEEauthorblockA{
        Faculty of Mechanical Engineering \\
        University of Victoria \\
        Victoria, BC, Canada \\
        mehranzoravar@uvic.ca
    }
    \and
    \IEEEauthorblockN{Shadi Alijani}
    \IEEEauthorblockA{
        Faculty of Electrical and Computer Engineering \\
        University of Victoria \\
        Victoria, BC, Canada \\
       shadialijani@uvic.ca
    }
    \and
    \IEEEauthorblockN{Homayoun Najjaran}
    \IEEEauthorblockA{
        Faculty of Mechanical,\\ Electrical, and Computer Engineering \\
        University of Victoria \\
        Victoria, BC, Canada \\
        najjaran@uvic.ca
    }
}

\begin{document}

\maketitle

\begin{abstract}
Exploring the trustworthiness of deep learning models is crucial, especially in critical domains such as medical imaging decision support systems. Conformal prediction has emerged as a rigorous means of providing deep learning models with reliable uncertainty estimates and safety guarantees. However, conformal prediction results face challenges due to the backbone model's struggles in domain-shifted scenarios, such as variations in different sources. To aim this challenge, this paper proposes a novel framework termed Conformal Ensemble of Vision Transformers (CE-ViTs) designed to enhance image classification performance by prioritizing domain adaptation and model robustness, while accounting for uncertainty. The proposed method leverages an ensemble of vision transformer models in the backbone, trained on diverse datasets including HAM10000, Dermofit, and Skin Cancer ISIC datasets. This ensemble learning approach, calibrated through the combined mentioned datasets, aims to enhance domain adaptation through conformal learning. Experimental results underscore that the framework achieves a high coverage rate of 90.38\%, representing an improvement of 9.95\% compared to the HAM10000 model. This indicates a strong likelihood that the prediction set includes the true label compared to singular models. Ensemble learning in CE-ViTs significantly improves conformal prediction performance, increasing the average prediction set size for challenging misclassified samples from 1.86 to 3.075.
\end{abstract}


%--------------------------- BODY ---------------------------

\section{Introduction}
In recent years, deep learning importance increased in a variety of research areas, most notably in medical image analysis with Convolutional Neural Networks (CNNs). Nevertheless, CNNs have limitations in dealing with long-term dependencies in view of localized receptive fields \cite{han2022survey}. 

Transformers, proposed by Vaswani et al. \cite{vaswani2017attention}, apply self-attention for feature extraction, and in consequence, revolutionized natural language processing with models such as BERT \cite{devlin2018bert} and GPT-3 \cite{brown2020language}. Motivated by such success, Dosovitskiy et al. \cite{dosovitskiy2020image} proposed Vision Transformers (ViTs) specifically for image classification, and these have seen improvements in computer vision-related performance. Recent works, such as DeepSkin \cite{gururaj2023deepskin}, SkinDistilViT \cite{lungu2023skindistilvit}, and YoTransViT \cite{saha2024yotransvit}, have emphasized the effectiveness of ViTs in skin lesion classification. In addition, Transformers have proven useful in use cases such as object detection and medical imaging \cite{alijani2024vision}. 

Ensemble learning aggregates several models for accuracy improvement and for easier domain adaptation, particularly in cases with multi-modal distributions \cite{zhou2021domain}. As important a metric accuracy is, its sole use in critical environments is not enough. Methods for Uncertainty Quantification (UQ) make a model more reliable through prediction confidence estimation \cite{fayyad2023empirical}. 

To tackle the problem with domain adaptation and robustness, in this work, Conformal Ensemble of Vision Transformers (CE-ViTs) is proposed, combining ViTs with Conformal Prediction (CP) for uncertainty estimation and utilizing ensemble learning techniques for generating robust and reliable forecasts in critical use cases.

\section{Background}
This section outlines key concepts related to Vision Transformers (ViTs), ensemble learning, and Conformal Prediction (CP), providing a foundation for understanding the methodologies used in this study.

\subsection{Vision Transformer in Domain Adaptation}
The Vision Transformer (ViT) utilizes transformer architectures for processing sequence consisting of image patches, including patch embedding, a transformer encoder, and a multi-layer perceptron (MLP) \cite{liu2021swin}. In its mechanism, images are partitioned into patches, subjected to linear transformations, and processed via self-attention processes allowing for extraction of global information \cite{han2022survey, vaswani2017attention, dosovitskiy2020image}. Unlike conventional convolutional neural networks (CNNs), ViTs exhibit high performance in dealing with shifts between domains, owed to its native global attention mechanism. Domain discrepancies have been addressed in current studies, such as Domain-Oriented Transformer (DOT) \cite{ma2022making}, Spectral UDA (SUDA) \cite{zhang2022spectral}, and CDTRANS \cite{li2022cdtrans}.

\begin{figure}[b]
    \centering
    \resizebox{0.75\linewidth}{!}{\includegraphics{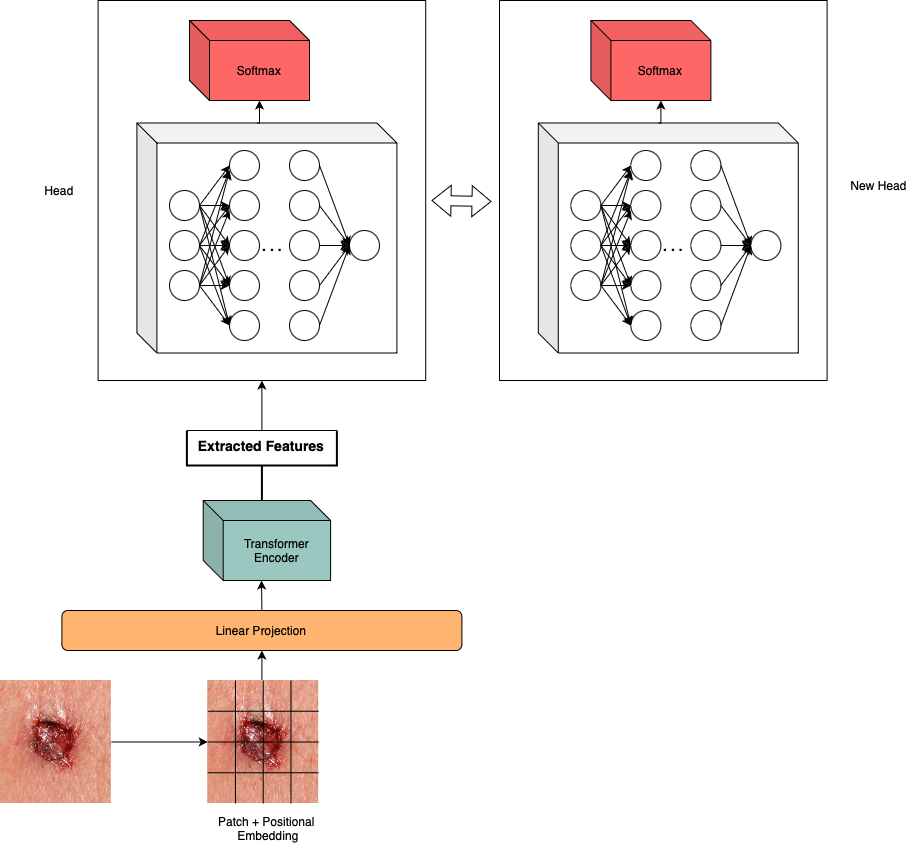}}
    \caption{A pre-trained ViT architecture with a new classifier head, fine-tuned for the skin lesion dataset.}
    \label{fig:onecol}
\end{figure}

\subsection{Ensemble Learning}
Ensemble learning improves predictive performance through combining several model output values with simple or weighted averaging techniques. This approach promotes robustness, especially when dealing with data drawn from disparate distributions, and is important for use in, for instance, skin lesion classification \cite{zhou2021domain}.

\subsection{Conformal Prediction}
Conformal Prediction (CP) provides uncertainty estimates by generating prediction sets that include the true label with a user-defined confidence level \cite{kasa2023empirically}. CP is model-agnostic, efficient, and reliable across different data types \cite{angelopoulos2020uncertainty}.

For a classification task, the SoftMax function $\sigma$ applied to model outputs generates conformity scores $S = \sigma f(X)$. Using a calibration dataset, the empirical quantile $\hat{q}$ is computed to form the prediction set:
\begin{equation}
    C(\textit{X}_{\text{test}}) = \{y: S(\textit{X}_{\text{test}}) < \hat{q} \}
\end{equation}
The prediction set includes classes where the SoftMax output is less than $1 - \hat{q}$. The validity of the prediction set is defined as \cite{angelopoulos2021gentle}:
\begin{equation}
1 - \alpha \leq P(Y_{\text{test}} \in C(X_{\text{test}})) \leq 1 - \alpha + \frac{1}{n+1}
\end{equation}
where $\alpha$ is the user-defined error rate. CP enhances model reliability, making it valuable for safety-critical applications like medical imaging \cite{fayyad2023empirical}.

\section{Methodology}
This section presents the proposed CE-ViTs framework, datasets used, and training configurations.

\subsection{Datasets}
To enhance model dependability in medical imaging, three datasets for skin lesions, namely DMF, HAM10000 \cite{Kaggle-skin-cancer-dataset}, and Skin Cancer ISIC \cite{Kaggle-skin-cancer-isic-dataset}, are utilized. There are seven classes in these datasets: Actinic Keratoses (akiec), Basal Cell Carcinoma (bcc), Benign Keratosis-like Lesions (bkl), Dermatofibroma (df), Melanoma (mel), Nevus (nv), and Vascular Lesions (vasc). To counteract class imbalance, several methodologies for data augmentation, including reflection, rotation, and scaling, have been utilized.

For conventional training, datasets have been partitioned into 70\% for training, 10\% for validation, and 20\% for testing. For Conformal Prediction (CP), distribution changed to 60\% for training, 10\% for validation, 20\% for calibration, and 10\% for testing. Having a combined test set helped in an unbiased evaluation of performance comparison.

\begin{figure*}[t]
    \centering
    \resizebox{0.6\linewidth}{!}{\includegraphics{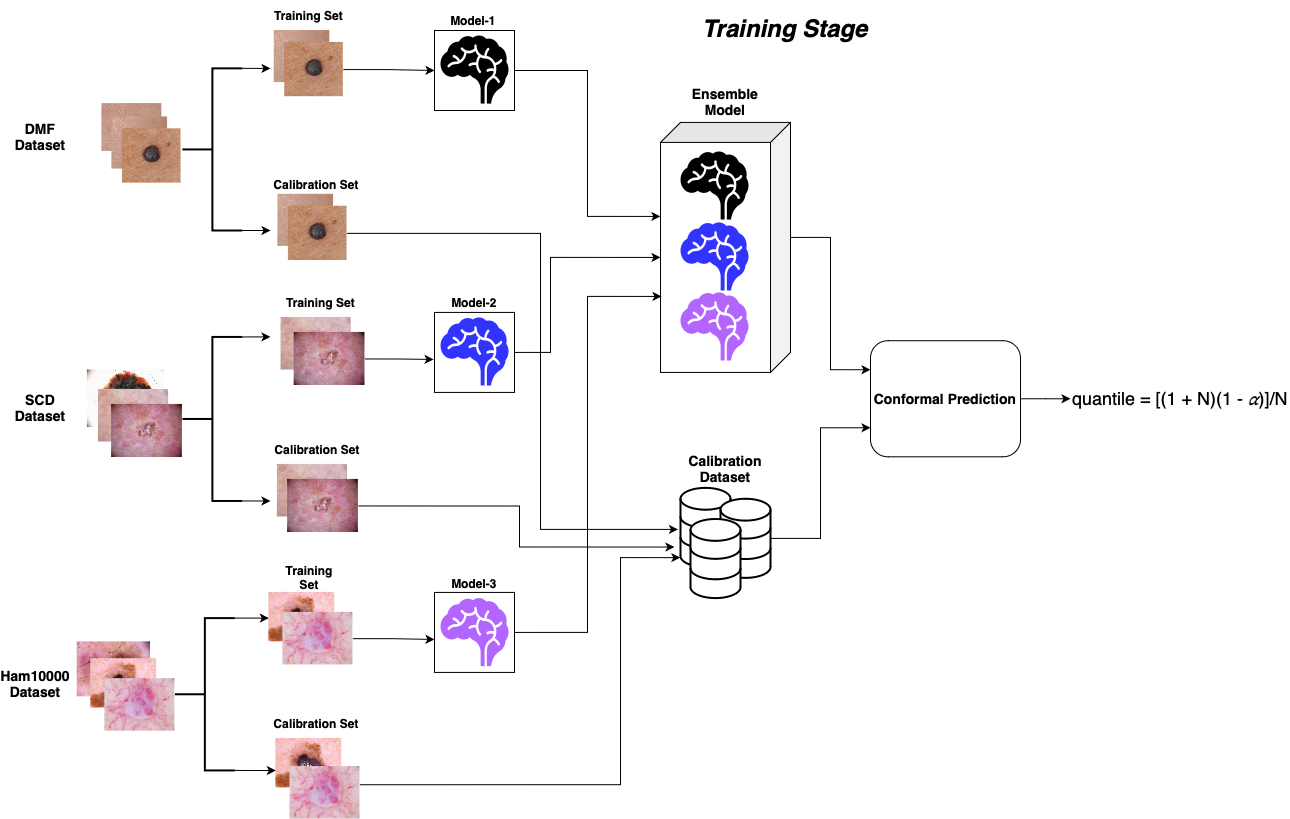}}
    \vspace{2mm}
    \resizebox{0.6\linewidth}{!}{\includegraphics{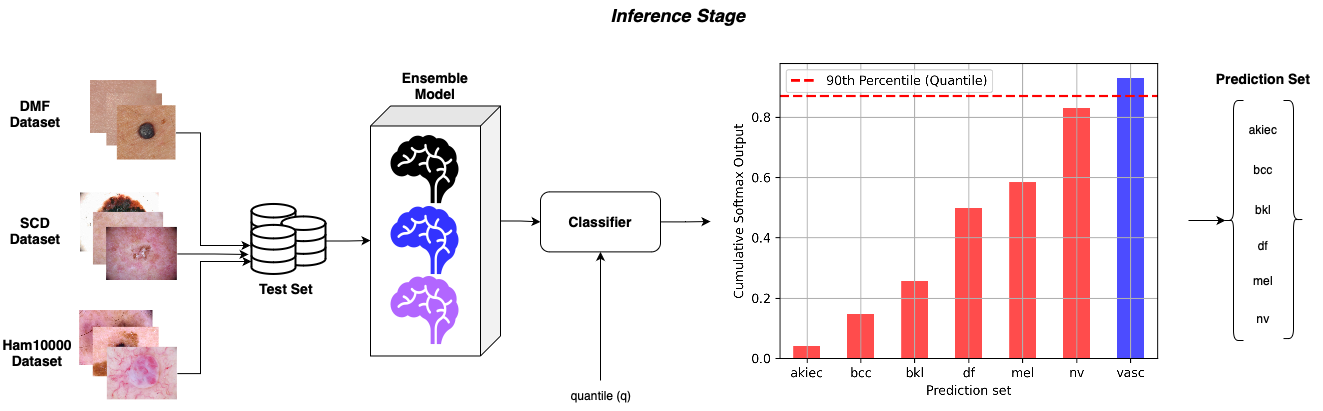}}
    \caption{The proposed CE-ViTs framework during training and inference stages.}
    \label{fig:framework}
\end{figure*}

\subsection{Conformal Ensemble of Vision Transformers (CE-ViTs)}
CE-ViTs employs ensemble learning on datasets with diverse distributions to improve domain adaptation \cite{zhou2021domain}. Models trained on different datasets are combined via simple averaging of their SoftMax outputs:
\begin{equation}
    output = Average(y_1, y_2, y_3)
\end{equation}
where \(y_i\) is the softmax output of model \(i\).

The CP is added as a post-processing module to generate reliable prediction sets. Calibration data integrates HAM10000, DMF, and ISIC, and in doing so, it strengthens robustness. As seen in Figure~\ref{fig:framework}, CP module employs the merged calibration data to estimate the quantile ($\hat{q}$), and in doing so, strengthens uncertainty quantification.

\subsection{Configuration and Training Setup}
A pre-trained Vision Transformer (ViT) model \cite{Pre-trained-model} was utilized, leveraging transfer learning with frozen backbone and replaced with an MLP head specifically for use in skin lesion classification. All models trained for 20 epochs with cross-entropy loss and optimized with stochastic gradient descent with momentum (SGDM) at a predefined learning rate of 0.0001.

The experiments were performed with a computer having an AMD Ryzen 9 3950X CPU and an NVIDIA Titan RTX GPU with 24 GB RAM, providing consistent performance in computationally intensive operations.

\section{Results and Discussion}
\label{sec:result and discussion}
\begin{figure}[t]
    \centering
    \resizebox{0.9\linewidth}{!}{\includegraphics{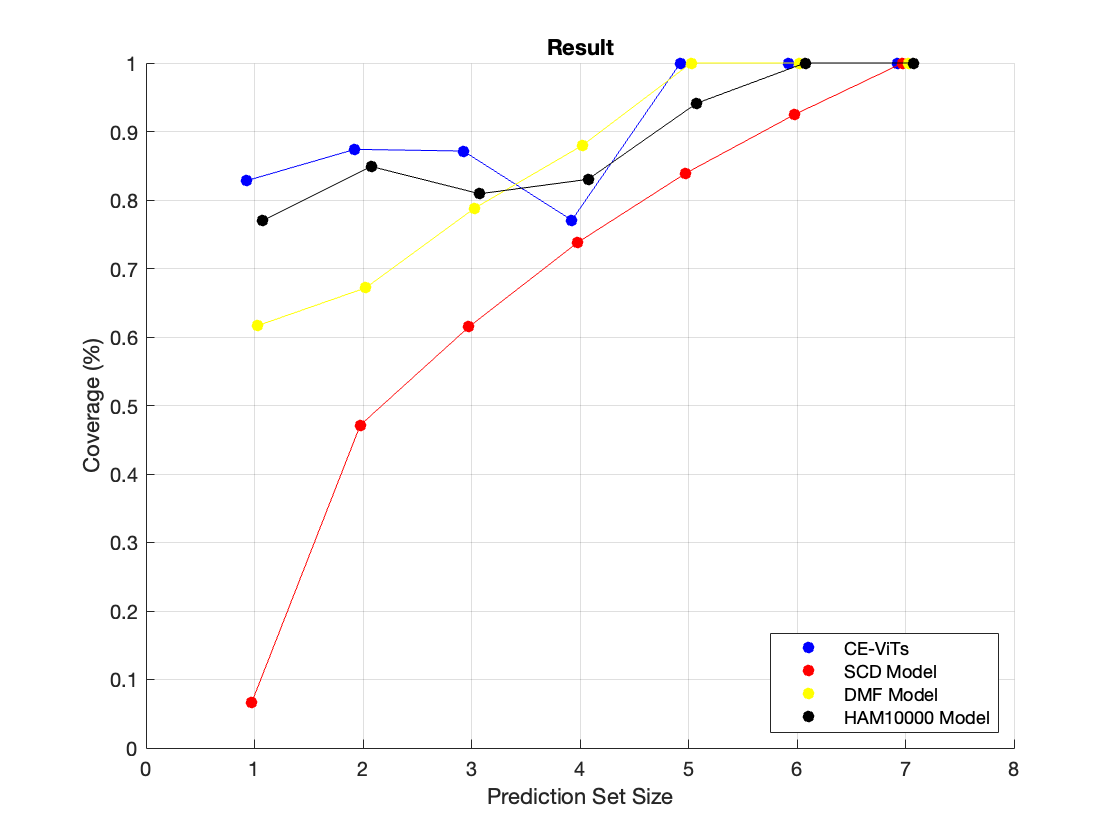}}
    \caption{Coverage and prediction set size correlation under the test dataset.}
    \label{fig:onecol}
\end{figure}

\begin{table*}[t]
  \centering
  {\small{
  \begin{tabular}{@{}lllllc@{}}
    \toprule
    Method &    Macro-Precision&Macro-Recall&Macro-F1Score& \(Accuracy^1 (\%)\)&\(Accuracy^2  (\%)\)\\
    \midrule
    Model-1 (HAM10000 Expert)&    0.6042&0.2725&0.2731& 72.80&66\\
    Model-2 (DMF Expert)&    0.3477&0.4914&0.3437& 53.72&48.51\\
    Model-3 (Skin Cancer ISIC Expert)&    0.2917&0.1727&0.1066& 49.28&15.67\\
 CE-ViTs&   0.4321&0.3784&0.2823& -&56.05\\
    \bottomrule
  \end{tabular}
  }}
  \caption{Model Metric\\
  \(^1\) This accuracy is calculated on the HAM10000 dataset for Model-1, the DMF dataset for Model-2, and the Skin Cancer ISIC dataset for Model-3.\\
    \(^2\) This accuracy is calculated based on the mixture of all the datasets.}
  \label{tab:example}
\end{table*}

The experimental results focus on two key dimensions: model evaluation and robustness, including accuracy, prediction set size, coverage, and uncertainty values.

\textbf{Accuracy.} In Table 1, performance statistics are aggregated, including accuracy, macro-recall, macro-precision, and macro-F1 Score. CE-ViTs shows increased accuracy compared to individual models for both the Skin Cancer ISIC and DMF datasets, suggesting increased generalization performance. Whereas HAM10000 expert model logs higher accuracy in its respective dataset, this is a result of its prioritization of less complex information. On the other hand, CE-ViTs, having been trained with a variety of distributions, shows increased robustness, specifically in real-world samples.

Aside from accuracy, in this work, uncertainty, particularly with regard to a variety of sources and distribution shifts, is evaluated to ensure reliable performance in real-world use cases.

\begin{figure*}[t]
    \centering
    % First row
    \begin{subfigure}{0.24\linewidth}
        \centering
        \includegraphics[width=\linewidth]{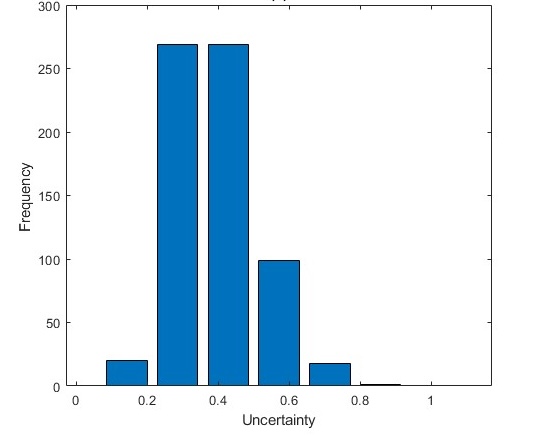}
        \caption{Correct predictions with CE-ViTs.}
        \label{fig:correct_ensemble}
    \end{subfigure}
    \hfill
    \begin{subfigure}{0.24\linewidth}
        \centering
        \includegraphics[width=\linewidth]{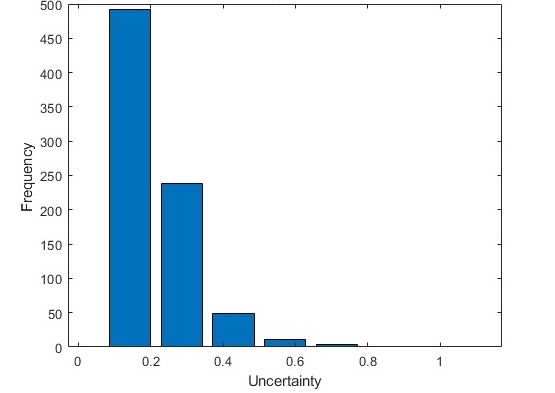}
        \caption{Correct predictions with HAM10000 expert model.}
        \label{fig:correct_ham}
    \end{subfigure}
    \hfill
    \begin{subfigure}{0.24\linewidth}
        \centering
        \includegraphics[width=\linewidth]{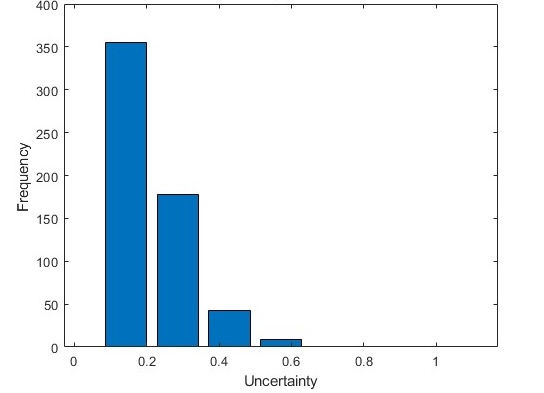}
        \caption{Correct predictions with DMF expert model.}
        \label{fig:correct_dmf}
    \end{subfigure}
    \hfill
    \begin{subfigure}{0.24\linewidth}
        \centering
        \includegraphics[width=\linewidth]{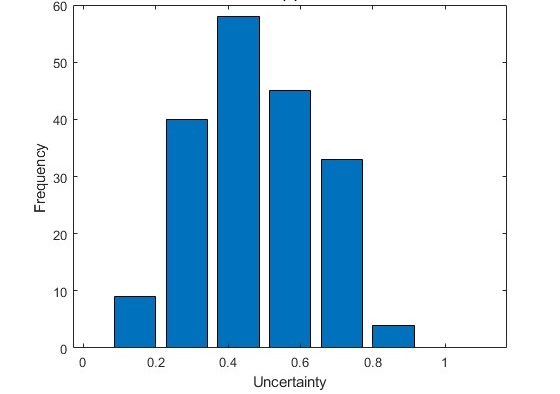}
        \caption{Correct predictions with Skin Cancer ISIC expert model.}
        \label{fig:correct_scd}
    \end{subfigure}

    % Second row
    \vspace{4mm}
    \begin{subfigure}{0.24\linewidth}
        \centering
        \includegraphics[width=\linewidth]{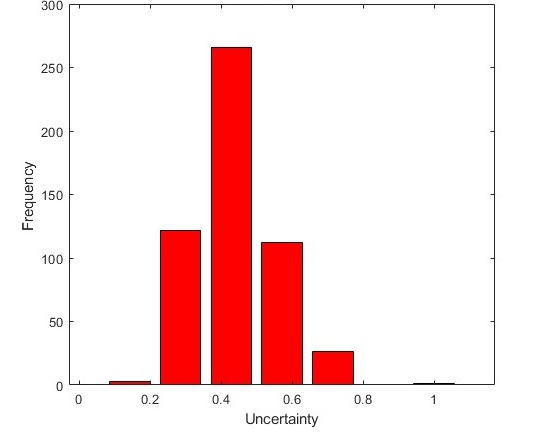}
        \caption{Incorrect predictions with CE-ViTs.}
        \label{fig:incorrect_ensemble}
    \end{subfigure}
    \hfill
    \begin{subfigure}{0.24\linewidth}
        \centering
        \includegraphics[width=\linewidth]{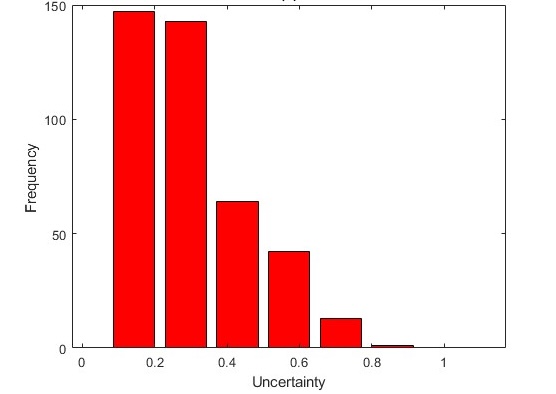}
        \caption{Incorrect predictions with HAM10000 expert model.}
        \label{fig:incorrect_ham}
    \end{subfigure}
    \hfill
    \begin{subfigure}{0.24\linewidth}
        \centering
        \includegraphics[width=\linewidth]{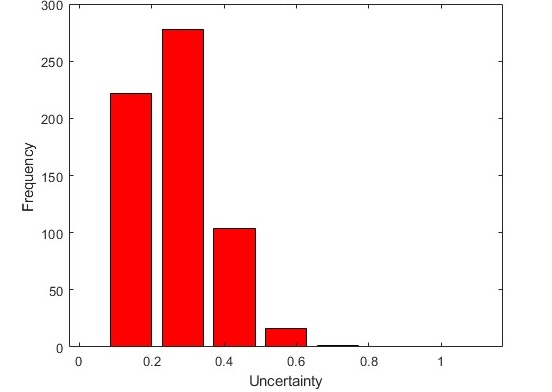}
        \caption{Incorrect predictions with DMF expert model.}
        \label{fig:incorrect_dmf}
    \end{subfigure}
    \hfill
    \begin{subfigure}{0.24\linewidth}
        \centering
        \includegraphics[width=\linewidth]{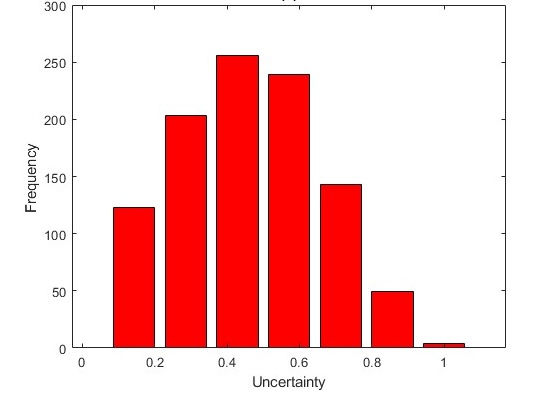}
        \caption{Incorrect predictions with Skin Cancer ISIC expert model.}
        \label{fig:incorrect_scd}
    \end{subfigure}
    \caption{The frequency and uncertainty value correlation for correct and incorrect predictions with different models.}
  \label{fig:short}
\end{figure*}    
\begin{table*}[t]
  \centering
  {\small{
  \begin{tabular}{@{}llll}
    \toprule
    Method &    \(C_{correct}\)&\(C_{incorrect}\)&Average\\
    \midrule
    Model-1 (HAM10000 Expert)&    1.49&2.10&1.795\\
    Model-2 (DMF Expert)&    1.50&1.86&1.68\\
    Model-3 (Skin Cancer ISIC Expert)&    3.34&3.23&3.285\\
 CE-ViTs&   2.74&3.075&2.9075\\
    \bottomrule
  \end{tabular}
  }}
  \caption{Comparison between the average prediction set size produced by the CP with using different models for correctly classified samples and wrongly classified samples. }
  \label{tab:example}
\end{table*}

**Prediction Set Size and Coverage**. Prediction set dimension is an important parameter for gauging predictive model utility in practice. A prediction set with an overly high dimension reduces the effectiveness of prediction at the expense of its specificity. On the other hand, coverage, understood as a proportion of the test set for which prediction set covers actual label, forms an important metric in its own right. Coverage describes a tradeoff between prediction set dimension and its capacity to cover actual labels. An average prediction set with a restricted dimension could not necessarily be beneficial, for it could miss actual label and, in consequence, threaten model trustfulness. In addition, a larger prediction set including actual label could face efficiency-related issues. Thus, a proper equilibrium between prediction set dimensions and coverage must be attained in order for both effectiveness and transparency of CP methodology to prevail.

Here, in terms of differentiation between accurately and misclassified images, and CP performance evaluation for two cases, distinguishing between accurately and misclassified images forms a first-step activity. Brief summary of observations concerning prediction set dimensions is included in Table 2.

The results reveal that individual model prediction sets have variable dimensions, depending on the model and its accuracy level. For correct prediction, prediction sets' dimensions range between 1.49 and 3.34. For its part, the CE-ViTs produce a value of 2.74. As much as a smaller prediction set for correct prediction is ideal, it is important to pay consideration to prediction set size in cases of incorrect prediction. Prediction sets must have a big enough size to cover actual labels. For incorrect prediction with the use of the CE-ViTs, prediction set size is 3.075, a value larger than both HAM10000 expert model and DMF expert model prediction sets' values. Apart from prediction set dimension, high coverage is important in terms of raising the chance of including actual labels in prediction sets.
The statistics for coverage calculated using a test dataset for individual expert models and for the ensemble model are presented below: 60.2\% for the Skin Cancer ISIC expert model, 66.42\% for DMF expert model, 80.43\% for HAM10000 expert model, and 90.38\% for the ensemble model. As such, these values confirm that use of CE-ViTs generates most ideal coverage, and therefore, a high chance that prediction sets will cover actual labels. In such a case, Figure 3 shows a plot between prediction set size and coverage for individual expert models and for use with the CE-ViTs. According to the information, for most prediction set sizes, a chance for actual labels to feature in prediction sets, represented through use of coverage, is high when using the CE-ViTs compared to individual expert models. As such, such an observation is significant in that it proves, in a conclusive manner, that use of CE-ViTs significantly raises model trustability.

In the summary of the performance with regard to prediction set size and its coverage, both HAM10000 expert model and DMF expert model have the least prediction set sizes for samples that have been predicted accurately. However, in consideration of the coverage statistics and as seen in Figure 3, it is beneficial to have a marginally larger prediction set size with a larger probability for including the actual label. For misclassified samples, the largest prediction set size is seen for the Skin Cancer ISIC expert model, but its coverage is relatively low. What these observations reveal is that a larger prediction set size for incorrect prediction, with a smaller prediction set size for correct prediction, is beneficial, particularly when consideration is taken for its impact on coverage. Consequently, observations reveal that the performance of CP is significantly boosted through its high coverage and proper prediction set size through CE-ViTs. As seen in Figure 3, its performance is remarkably high for prediction set sizes 1, 2, and 3, and these are logical options for datasets with 7 classes.

\textbf{Uncertainty Value}. Uncertainty value describes the level of unpredictability of forecasts produced by a model. It is an important metric for gauging the effectiveness of classification processes (CP), particularly in critical industries such as medical and healthcare datasets. It is important to understand the level of uncertainty in a prediction in order to maintain accuracy and dependability in output. Figure 4 shows graphical representations of uncertainty values for individual models and for CE-ViTs.

For correct samples, Figures 4 (b) and (c) illustrate that uncertainty values for HAM10000 expert model and DMF expert model are relatively low. On the other hand, graph (d) shows that correct classification performed by the Skin Cancer ISIC expert model comes with exceedingly high uncertainty, and that is not satisfactory. In contrast, graph (a) reveals that CE-ViTs have low to intermediate uncertainty values, and that is satisfactory in view of high coverage rate.

For incorrect samples, Figure 4 graph (e) reveals satisfactory performance of CE-ViTs in dealing with incorrect forecasts. The minimum uncertainty value for incorrect forecasts tends towards 0, and that signifies that such forecasts consistently have high uncertainty, a key feature in cases of misclassification. On the other hand, graphs (f) and (g) illustrate low uncertainty values for incorrect forecasts, and that signifies that CP modules in such cases lack effectiveness. CP's key objective is to represent high uncertainty when a prediction is incorrect in a move to counteract overconfidence errors. As such, CE-ViTs vastly outdoes its counterpart models, particularly in cases of misclassified samples. It pairs incorrect forecasts with high uncertainty, and that helps in evading overconfidence mistakes. That is a significant improvement in uncertainty estimation, particularly in CP frameworks.

\section{Conclusion}
\label{sec:conclusion}

This research presents the CE-ViTs framework to address domain adaptation challenges and uncertainty in medical imaging. By integrating vision transformers trained on heterogeneous datasets (HAM10000, Dermofit, and Skin Cancer ISIC) through ensemble learning, CE-ViTs enhances prediction reliability and robustness.

Our experiments show that CE-ViTs achieves over 90\% coverage, ensuring the true label is included more frequently than with individual models. Additionally, it effectively handles difficult classification cases with improved prediction set sizes for misclassified samples.

The results highlight the benefits of combining ensemble learning with conformal prediction to improve performance across diverse data distributions. In future work, we will explore weighted ensemble learning for better calibration and investigate integrating conformal prediction directly into the training process to further boost robustness and accuracy in medical imaging applications.

%--------------------------- REFERENCES ---------------------------

\bibliographystyle{IEEEtran}
\bibliography{references}

\end{document}